\documentclass[conference]{IEEEtran}
\IEEEoverridecommandlockouts

\usepackage{cite}
\usepackage{url}
\usepackage{amsmath,amssymb,amsfonts}
\usepackage{algorithmic}
\usepackage{graphicx}
\usepackage{textcomp}
\usepackage{xcolor}
\usepackage{multirow}
\usepackage{float}
\usepackage{subfig}
\usepackage{makecell} 

\def\BibTeX{{\rm B\kern-.05em{\sc i\kern-.025em b}\kern-.08em
    T\kern-.1667em\lower.7ex\hbox{E}\kern-.125emX}}

\begin{document}

\title{BEACON: Behavioral Malware Classification with Large Language Model Embeddings \\and Deep Learning}

\author{
\IEEEauthorblockN{Wadduwage Shanika Perera}
\IEEEauthorblockA{\textit{Department of Computer Science} \\
\textit{Sam Houston State University}\\
Huntsville, TX 77341, USA \\
}
\and
\IEEEauthorblockN{Haodi Jiang$^*$\thanks{$^*$Corresponding author: haodi.jiang@shsu.edu}}
\IEEEauthorblockA{\textit{Department of Computer Science} \\
\textit{Sam Houston State University}\\
Huntsville, TX 77341, USA}
}

\maketitle

\begin{abstract}
Malware is becoming increasingly complex and widespread, making it essential to develop more effective and timely detection methods. Traditional static analysis often fails to defend against modern threats that employ code obfuscation, polymorphism, and other evasion techniques. In contrast, behavioral malware detection, which monitors runtime activities, provides a more reliable and context-aware solution. 
In this work, we propose BEACON, a novel deep learning framework that leverages large language models (LLMs) to generate dense, contextual embeddings from raw sandbox-generated behavior reports. 
These embeddings capture semantic and structural patterns of each sample and are processed by a one-dimensional convolutional neural network (1D CNN) for multi-class malware classification.
Evaluated on the Avast-CTU Public CAPE Dataset, our framework consistently outperforms existing methods, highlighting the effectiveness of LLM-based behavioral embeddings and the overall design of BEACON for robust malware classification.
\end{abstract}


\section{Introduction}
Malware evolution presents persistent challenges to cybersecurity. These threats are primary causes of system compromise and operational disruption, underscoring the need for more effective detection methods. Reliable identification of malware is important to initiate rapid mitigation measures, contain threats, and prevent widespread system compromise. Traditional signature-based malware detection is a widely used and straightforward method that relies on analyzing malware samples to extract unique signatures and identify malicious files without executing them \cite{Liu2011IWCDM}. Although it offers computational efficiency and faster processing, its effectiveness is limited when dealing with encrypted samples and is proven inadequate against modern malware, which constantly mutates and uses obfuscation and runtime packing to evade identification \cite{bosansky2022avastdataset}. These limitations have led to a shift toward dynamic malware analysis, which executes malware to observe its behavior, revealing hidden or environment-dependent features, even analyzing obfuscated or packed code that static methods often miss \cite{Yanfang2018ACM}.

Behavioral malware analysis, a key dynamic analysis technique, monitors and logs the actions of a malware program during its execution in a controlled environment, using the collected behavioral information as the foundation for detection, thereby reducing dependence on static characteristics. This makes it resistant to evasion techniques such as shell code injection, polymorphism, and metamorphism \cite{Liu2011IWCDM}. Malware behavior is often more distinctive than its static features, as core behavioral patterns are retained even as it mutates, making behavior-based detection effective against previously unseen or obfuscated threats. \cite{bosansky2022avastdataset}. Although this approach requires more computational resources and time, it plays a crucial role in complementing static analysis and enhancing overall malware detection strategies. However, the growing volume and complexity of malware call for scalable and automated behavioral malware detection methods.

Machine learning (ML) and deep learning (DL) techniques have become essential tools in the malware analysis field, supporting both the automation of feature extraction from behavioral reports and the development of effective malware classification systems \cite{Daniel2022usenixsecurity22, Jiang2018ICMLA, M2023100529}.
These approaches rely on well-structured feature representations that capture the distinguishing characteristics of malicious behavior. 
In particular, ML and DL models have been widely applied to learn discriminative patterns from raw or engineered features, enabling the detection and classification of malware across diverse families and obfuscation strategies \cite{raff2017malwaredetectioneatingexe, Jiang_Polsani_Liu_2024}.
Given the notable differences in the contextual execution behaviors of malware and benign software (goodware) \cite{AMER2020101760}, incorporating contextual relationships into feature representations is essential for improving detection accuracy. Large language models (LLMs), with their transformer-based architectures and self-attention mechanisms, have demonstrated remarkable success in capturing deep contextual and semantic relationships in natural language \cite{tao2024llmseffectiveembeddingmodels}. LLMs have demonstrated considerable potential in broader cybersecurity domains, including security and privacy \cite{yao2024survey}.

In this work, we propose Behavioral Embedding-Aware Convolutional Neural Network (BEACON), a deep learning framework that utilizes large language model (LLM)-generated embeddings as feature representations for malware classification.
Rather than relying on manual feature engineering, BEACON extracts semantic embeddings from textual behavioral reports using Google's textembedding-gecko@003 model, part of the Gemini family of LLMs \cite{googletextembeddingmodel}. This approach streamlines the feature extraction process while enhancing the contextual depth of the representations. By leveraging the ability of LLMs to produce nuanced and semantically rich embeddings that capture hierarchical and temporal dependencies, BEACON  enables robust malware representations without the need for additional model training. 
To perform malware family classification, the framework incorporates a one-dimensional convolutional neural network (1D CNN) that processes the LLM-derived embeddings.

The primary contribution of our work can be summarized as follows. 
\begin{itemize}
\item We introduce BEACON, a deep learning framework for malware classification that leverages a pre-trained LLM to extract dense contextual embeddings from raw behavioral reports, bypassing traditional hierarchical feature engineering.
\item Within the BEACON framework, we incorporate a one-dimensional convolutional neural network (1D CNN) to classify malware families based on the LLM-derived embeddings. We evaluate its performance against baseline models, including SVM, MLP, and LSTM, all using identical input representations, to demonstrate its effectiveness in behavior-based malware detection.
\item BEACON provides a compact and semantically rich vector representation of complex behavioral data, enabling efficient subsequent classification and offering a flexible foundation for future advancements in LLM-based malware analysis.
\end{itemize}

The remainder of this paper is organized as follows. Section II reviews related work on malware classification using various approaches. Section III introduces the dataset used in this study. Section IV details the methodology, including data pre-processing, embedding extraction, and model training. Section V presents the experimental results and compares them with existing methods. Finally, Section VI concludes the paper and outlines directions for future work.

\section{Related Works}

This section reviews the literature on malware detection and classification approaches based on behavioral malware analysis. Early work by Wu et al. \cite{Liu2011IWCDM} demonstrated the effectiveness of extracting malware behavior characteristics, such as subprocess creation, kernel driver loading, thread injection, self-deletion, and auto-start mechanisms, during malware execution. These features enabled the detection of previously unseen malware and showed robustness against common evasion strategies, including shellcode obfuscation, polymorphism, and metamorphism.
 
Bosansky et al. \cite{bosansky2022avastdataset} introduced the Avast-CTU Public CAPE Dataset, which provides detailed behavioral reports in JSON format, including accessed files, registry keys, mutexes, and API calls, alongside static features extracted from PE files.
Their study showed that detection models trained on full reports, combining behavioral and static data, outperformed those relying solely on static features. 
In particular, static-only models failed to generalize under concept drift, especially for malware families such as Emotet and Qakbot. Their hierarchical multi-instance learning (HMIL) framework preserved the hierarchical structure of JSON reports but struggled to capture deeper contextual relationships and was limited by high-dimensional data.
Ilic et al. \cite{ilicGoingAPICalls2024} further highlighted the value of diverse behavioral features by creating a dataset that captured file system changes, registry modifications, and network activity alongside API calls. Their results demonstrated that models trained on complete sandbox reports consistently outperformed those using only API call sequences, underscoring the importance of comprehensive behavioral context for robust malware detection.

Early approaches to feature representation primarily relied on manual techniques, such as feature hashing \cite{Zhang2020AAAI}, which enabled efficient extraction but often suffered from hash collisions and a lack of semantic clarity. 
Subsequent efforts employed word embedding models, such as Word2Vec \cite{kale2021Word2Vec}, to encode behavioral strings; however, these models assign fixed vector representations to tokens irrespective of context, thereby limiting their expressiveness in dynamic behavioral logs.
Regeciova et al. \cite{Regeciova2023GenRex} introduced a method based on regular expression patterns and behavioral YARA rules to construct a common pattern set. However, their dependence on predefined patterns limits adaptability to obfuscated or novel malware variants.

To overcome these limitations, more advanced neural architectures have been developed for representation learning. Neurlux \cite{jindal2019neurlux} tokenized behavioral report content and learned trainable word embeddings, combining CNNs for local pattern extraction, BiLSTMs for sequence modeling, and hierarchical attention to weigh important tokens. While this architecture eliminates the need for manual feature engineering, it is sensitive to inconsistent token sequences and susceptible to evasion via noise injection. Nebula \cite{Trizna2024Nebula}, a transformer-based dynamic malware analyzer, applied Byte Pair Encoding (BPE) to behavioral logs and utilized learned positional embeddings to capture long-range dependencies. Although effective in contextual modeling, the tokenization process sometimes fragments domain-specific terms, reducing semantic fidelity.
Other approaches have investigated structural representations to encode malware behavior. MalDetConv \cite{maniriho2022MalDetConv} used NLP-style encoders to transform API call sequences into dense vector representations, enabling automated feature extraction in behavior-based malware detection frameworks. Xiao et al. \cite{xiaoMalwareDetectionBased2019} proposed a behavior-based deep learning framework that uses stacked autoencoders to automatically extract high-level semantic features from behavior graphs constructed from Windows API calls. Although behavior graphs can preserve structural relationships between actions, they often fail to model the temporal and contextual dependencies necessary for detecting sophisticated or evasive malware.

Recent work has turned to large language models (LLMs) for malware classification and analysis tasks, leveraging their ability to generate semantically rich and context-aware embeddings. Bayer et al. \cite{bayer2022cysecbertdomainadaptedlanguagemodel} introduced CySecBERT, a BERT model fine-tuned on a cybersecurity-specific domain.
Ghourabi \cite{ghourabi2022security} applied pre-trained BERT to behavioral text and demonstrated improved semantic representation for malware detection. 
Sánchez et al. \cite{sánchez2024transferlearninglarge} developed a malware classification framework using pre-trained LLMs, including LLMs BERT, DistilBERT, GPT-2, BigBird, Longformer, and Mistral, based on system calls, and found that transformer-generated contextual embeddings consistently outperformed conventional representations\cite{sánchez2024transferlearninglarge}.
Marais et al. \cite{marais2025semantic} applied LLMs to summarize PE file reports and classified the generated summaries using a BERT model, improving both interpretability and classification performance.
Li et al. \cite{Li2023Efficient} transformed PE files into grayscale image representations and used LLaMA-7b to extract structural and semantic patterns from the visualized binary data. 
Qian et al. \cite{qian2025lamd} developed a framework for Android malware detection using GPT-4o-min to derive context-driven predictions and explanations, enhancing the understanding of the malware’s operation. 
 
Despite these advances, the direct use of LLM-generated contextual embeddings for malware family classification from raw behavioral reports remains underexplored. 
Existing methods still face challenges in fully capturing rich contextual relationships within behavioral data. 
To address this gap, we propose a deep learning framework that extracts dense, context-aware embeddings directly from sandbox-generated logs using a state-of-the-art large language model. These embeddings are subsequently processed by a one-dimensional convolutional neural network, enabling robust malware family classification through semantically rich and temporally informed representations.

\section{Dataset}

The dataset used in this study, the Avast-CTU Public CAPE Dataset, was first introduced by Bosansky et al \cite{bosansky2022avastdataset}. It is a rich behavioral dataset of 48,976 malware behavioral reports in JSON format classified into 10 different malware families, as shown in Table \ref{tab:malware_family_distr}. The malware activity reports were primarily collected between 2017 and 2019, providing long-term data distribution changes such as concept drift. Each sample is associated with metadata, including malware family, malware type, detection date, and the report’s SHA-256 hash for identification.

\begin{table}[ht]
\caption{Distribution of Malware Samples by Type and Family}
\label{tab:malware_family_distr}
\centering
\renewcommand{\arraystretch}{1.5} 
\begin{tabular}{|c|c|c|}
\hline
\textbf{Malware Type} & \textbf{Malware Family} & \textbf{Number of Instances} \\ 
\hline
\multirow{5}{*}{banker} & Emotet & 14429 \\ 
\cline{2-3}
& Qakbot & 4895 \\ 
\cline{2-3}
& Trickbot & 4202 \\
\cline{2-3}
& Ursnif & 1343 \\
\cline{2-3}
& Zeus & 2594 \\
\hline
coinminer & HarHar & 655 \\ 
\hline
keylogger & njRAT & 29 \\ 
\hline
pws & Lokibot & 4191 \\ 
\hline
rat & njRAT & 3343 \\ 
\hline
\multirow{2}{*}{trojan} & Adload & 704 \\ 
\cline{2-3}
& Swisyn & 12591 \\
\hline
\end{tabular}
\end{table}

The malware samples were executed and analyzed using the CAPEv2 sandbox, an open-source dynamic analysis framework that extends the capabilities of the original Cuckoo sandbox \cite{bosansky2022avastdataset}. CAPEv2 improves the monitoring by capturing payloads, unpacking binaries, and detecting malware behaviors through features such as YARA (Yet Another Ridiculous Acronym) signature matching \cite{yara}. For each executed sample, the dataset provides a full JSON report generated by CAPEv2, containing comprehensive behavioral information, including process activity, API calls, dropped files, and detection results. To address potential issues with label leakage and enormous report sizes (some exceeding 800 MB), a set of reduced JSON reports was also created. These reduced reports focus on two main components: a behavioral summary and static PE file information. This reduction preserves essential features for machine learning tasks while improving efficiency and mitigating risks related to data leakage. We used reduced reports in our study to focus on essential static and behavioral characteristics, while excluding information such as YARA-based detections that could artificially boost classification performance. In addition, this approach significantly reduced the size of the input data, making model training more efficient and scalable without compromising the richness of behavioral analysis.

\section{Methodology}

\begin{figure*}[ht]
\centering
\includegraphics[width=\linewidth]{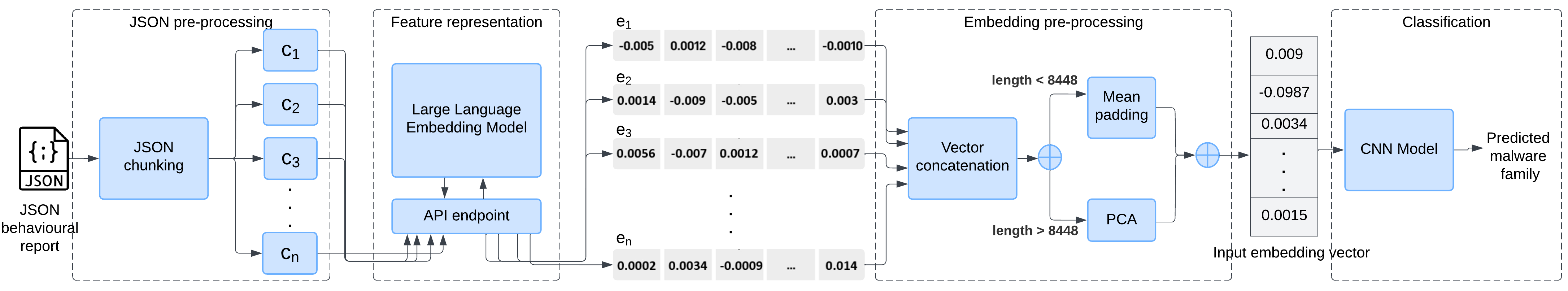}
\caption{BEACON: The proposed framework for behavioural malware classification.}
\label{fig:proposed_framework}
\end{figure*}

This section outlines the methodology used in this study. Figure \ref{fig:proposed_framework} illustrates the overall workflow we propose for malware classification based on behavioral malware reports. The workflow comprises data pre-processing, feature representation, and model training, each of which is described in detail below.

\subsection{JSON Pre-processing}

We used the textembedding-gecko@003 model through the Google Cloud’s Gemini API, which has a 2,048 token limit per input text \cite{googletextembeddingmodel}. The majority of the reduced JSON malware report produced by the CAPEv2 sandbox exceeded this limit; therefore, each JSON file was first serialized into a plain text format preserving its hierarchical structure, and then divided into $n$ chunks $c_1, c_2, \dots, c_n$ (Figure~\ref{fig:proposed_framework}), where $n$ varies depending on the file size, before being passed to the embedding model. The files were chunked based on the approximation that one token corresponds to about four characters ($1\ \text{token} \approx 4\ \text{characters}$) for Gemini models \cite{googletokens}.

To chunk the JSON files while preserving their structural integrity, we used the RecursiveJsonSplitter from LangChain \cite{langchainjsonsplitter}. Unlike traditional text splitting methods that split purely on whitespace or simple character counts, the RecursiveJsonSplitter employs a structure-aware approach specifically tailored for nested JSON data. When a JSON object exceeds the specified maximum chunk size, the splitter recursively breaks it down into smaller, manageable pieces while maintaining the nested relationships between elements. Additionally, it converts lists in JSON structures into dictionaries for improved segmentation.
This method of preserving hierarchical relationships within the JSON files ensures the embeddings accurately capture the context and nuanced behavioral patterns of malware, which is essential for maintaining high classification fidelity across diverse malware families. Each chunk created was subsequently passed to the LLM embedding model.

\subsection{Feature Representation}

We used dense contextual embeddings derived from raw JSON behavioral reports to represent features. Dense text embeddings are high-dimensional numerical vector representations of natural language that capture semantic relationships. We used the textembedding-gecko@003 model to generate embedding vectors. This model is designed to generate dense, high-dimensional embeddings from raw text input, capturing rich semantic and contextual information based on the surrounding text, enabling a more expressive and informative feature representation. Each chunk of an input JSON file, produced during the JSON pre-processing stage, was passed through the embedding model to produce embedding vectors $e_1, e_2, \dots, e_n$, each being a fixed-length vector of size 768 (Figure~\ref{fig:proposed_framework}). The resulting dense contextual embeddings $e_1, e_2, \dots, e_n$ that correspond to an input file represent the semantic and behavioral pattern features present in that particular malware report, forming the input features for our classification models.

\subsection{Embedding Pre-processing}  

\begin{figure}[h]
\centering
\includegraphics[width=0.96\linewidth]{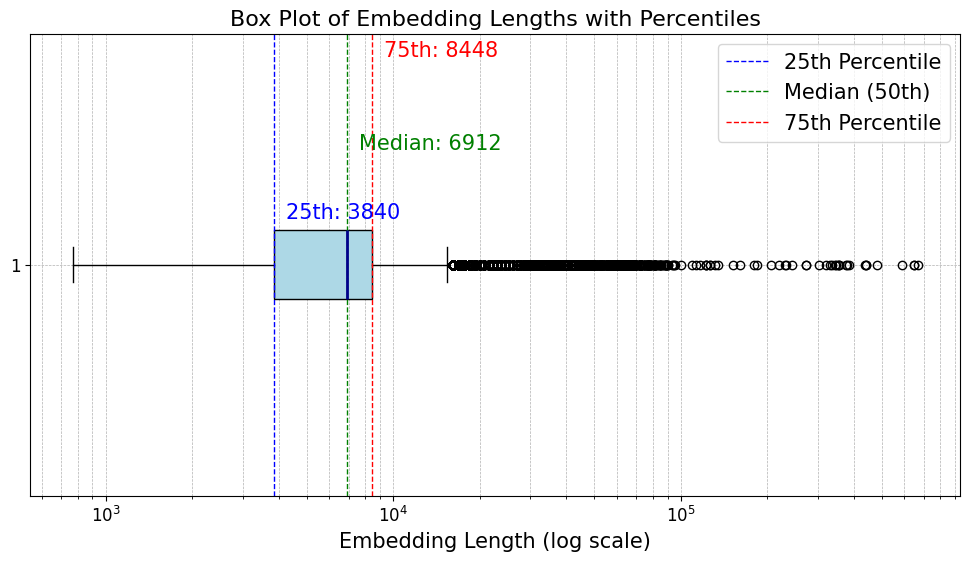}
\caption{Distribution of embedding lengths of Avast-CTU Dataset in log scale.}
\label{fig:embedding_lengths}
\end{figure}

After generating the embedding vectors for each chunk of an input file, the vectors were concatenated to form a single 1-D vector for the entire file (Figure~\ref{fig:proposed_framework}). The resulting embeddings obtained from concatenation varied in length due to the diverse size distribution of the original JSON reports. The embedding length distribution presented in Figure \ref{fig:embedding_lengths} demonstrates that most samples have relatively short lengths, whereas only a few samples exhibit exceptionally large lengths. 

To ensure all input embeddings have a consistent size, Principal Component Analysis (PCA) was applied to the lengthy vectors to reduce their dimensionality to the 75th percentile length of all embeddings (8,448) and padded the remaining shorter vectors with their respective mean values, resulting in all vectors having a uniform fixed length (Figure \ref{fig:proposed_framework}). The 75th percentile was chosen as it provides a balance between minimizing information loss from longer embeddings and avoiding excessive padding for shorter ones. 
Dimensionality reduction and padding ensure consistent input sizes for stable CNN training.

\subsection{Model Training}

\paragraph{One-Dimensional Convolutional Neural Network}
As the core classifier within the BEACON framework, we design a customized one-dimensional convolutional neural network (1D CNN) to perform malware family classification based on LLM-generated embeddings, as shown in Figure~\ref{fig:proposed_framework}.
The architecture is designed to automatically learn hierarchical representations of malware behavior from the input embeddings through a series of stacked 1D convolutional layers, each followed by batch normalization and ReLU activation. Max-pooling layers are applied after each convolutional block to
reduce dimensionality while preserving critical features. The resulting feature maps are flattened and passed through a dense layer with a dropout rate of 0.5, followed by a final fully connected classification layer. 
This model component enables BEACON to capture high-level semantic patterns directly from raw embeddings without relying on manually engineered features. Table \ref{table:summary_cnn_archtecture} outlines the complete architecture. 

\begin{table}[H]
    \caption{Architecture of the 1D CNN model in BEACON}
    \centering
    \renewcommand{\arraystretch}{1.5} 
    \begin{tabular}{|c|c|c|c|}
        \hline
        \textbf{Layer} & \textbf{\makecell{Parameters \\ and Activations}} &  \textbf{\makecell{Feature \\ Map}} & \textbf{\makecell{Output \\ Shape} } \\
        \hline 
         Input & - & 1 & 1x8448 \\ 
        \hline 
        Conv1D & \makecell{kernel=5, stride=1, \\ padding=2, activation=relu} & 32 & 32x8448  \\ 
        \hline 
        BatchNorm1D & momentum=0.1 & 32 & 32x8448 \\
        \hline 
        MaxPool1D & kernel=2, stride=2 & 32 & 32x4224 \\
        \hline 
        Conv1D & \makecell{kernel=5, stride=1, \\ padding=2, activation=relu} & 64 & 64x4224 \\
        \hline 
        BatchNorm1D & momentum=0.1 & 64 & 64x4224 \\
        \hline 
        MaxPool1D & kernel=2, stride=2 & 64 & 64x2112 \\
        \hline 
        Conv1D & \makecell{kernel=5, stride=1, \\ padding=2, activation=relu} & 128 & 128x2112 \\
        \hline 
        BatchNorm1D & momentum=0.1 & 128 & 128x2112 \\
        \hline 
        MaxPool1D & kernel=2, stride=2 & 128 & 128x1056 \\
        \hline 
        Flatten & - & - & 135,168 \\
        \hline
        Dense & \makecell{in=135168, out=512 \\ activation=relu} & - & 512\\
        \hline
        Dropout & rate = 0.5 & - & 512\\
        \hline
        Dense (Output) & \makecell{in=512, out=10, \\ activation=softmax} & - & 10\\
        \hline
    \end{tabular} 
    \label{table:summary_cnn_archtecture}
\end{table}

\paragraph{Baseline Models for Comparison}

To assess the classification performance of BEACON, we compare its core 1D CNN component against several baseline models. All baseline classifiers are trained using the same LLM-generated embeddings to ensure a fair and consistent evaluation setting.

\textit{Support Vector Machine (SVM):}  
A classical SVM classifier with a radial basis function (RBF) kernel was implemented. The RBF kernel enables the model to capture non-linear decision boundaries within the embedding space. Hyperparameters were optimized using grid search, yielding \( C = 0.01 \) and \( \gamma = 100 \).

\textit{Multilayer Perceptron (MLP):}  
The MLP model was constructed following the architecture reported in \cite{bosansky2022avastdataset}. The model consists of a single hidden layer with 32 neurons and a ReLU activation function, positioned between an input layer of size 8,448 and an output layer of size 10. This serves as a simple yet informative baseline for assessing the effectiveness of the embedding representations and downstream classifiers.

\textit{Bidirectional LSTM:}  
We implemented a two-layer bidirectional long short-term memory (LSTM) network to capture temporal dependencies within the sequential embedding vectors. This model processes the sequence in both forward and backward directions, enabling the capture of context from both preceding and succeeding elements. The detailed architecture is shown in Table~\ref{table:summary_lstm_archtecture}.

\begin{table}[H]
    \caption{Architecture of the Bidirectional LSTM}
    \centering
    \renewcommand{\arraystretch}{1.5} 
    \begin{tabular}{|c|c|c|}
        \hline
        \textbf{Layer/Step} & \makecell{\textbf{Parameters/} \\ \textbf{Output Representation}} & \makecell{\textbf{Output Shape} \\ \textbf{(x batch\_size)}} \\ 
        \hline 
        Input & - & 1×8448 \\ 
        \hline 
        LSTM & \makecell{input size=8448, hidden dim=128, \\ layers=2, bidirectional=True} & 1×256 \\ 
        \hline 
        \makecell{Last Time \\ Step Slice} & \makecell{all features from \\ the last time step} & 256 \\
        \hline 
        Dense (Output) & in=256, out=10 & 10 \\
        \hline
    \end{tabular} 
    \label{table:summary_lstm_archtecture}
\end{table}

\paragraph{Training Configuration}  
All models were trained to classify samples into 10 malware families using the same dataset and embedded vectors. 
Neural network models, including the core 1D CNN classifier, as well as the baseline MLP and LSTM models, were optimized using the Adam optimizer with a learning rate of 0.0001 and categorical cross-entropy loss. The 1D CNN and MLP models were trained with a batch size of 64, while the LSTM model used a batch size of 32. All models were trained for 200 epochs and used a softmax activation function in the output layer.

Hyperparameters for all four models were selected via grid search to ensure fair comparison \cite{2018scikitlearn}. These configurations enabled a systematic evaluation of BEACON against representative classical and deep learning approaches under consistent experimental conditions.

\section{Experiments and Results}

In this section, we present the experimental setup and evaluate the performance of the proposed malware classification framework. 
Following the approach described in the Avast-CTU Public CAPEv2 dataset paper \cite{bosansky2022avastdataset}, we adopted a temporal split of the dataset, allocating 76\% of the samples for training and the remaining 24\% for testing. To ensure a robust and unbiased evaluation, we applied 5-fold cross-validation on the training set. This approach ensures that each data point is used for both training and validation across different folds, providing a more stable performance estimate while reducing the risk of overfitting.

\subsection{Evaluation Metrics}

We adopt five metrics to evaluate the performance of the classification models: accuracy, precision, recall, F1 score, and the area under the precision–recall curve (AUPRC).
These metrics are computed using true positives (TP), true negatives (TN), false positives (FP), and false negatives (FN). TP refers to the number of instances of a specific malware family that are correctly predicted. FP refers to the number of instances where the model incorrectly classifies a sample as a specific malware family, but it belongs to a different class or is benign. TN refers to the number of correctly identified samples that do not belong to the predicted malware class. FN refers to the number of instances where the model incorrectly classifies a sample as benign or as a different malware family when it belongs to the predicted malware class.
In this study, we report weighted averages of precision, recall, and F1-score to account for class imbalance in our multi-class setting. 
AUPRC is included as a threshold-independent measure that is particularly informative under class imbalance.

\textbf{Accuracy (ACC):}  ACC measures the proportion of correctly classified samples out of the total number of samples. It provides an overall indicator of the model’s classification performance.

\begin{equation}
\text{ACC} = \frac{TP + TN}{TP + TN + FP + FN}
\label{eq:accuracy}
\end{equation}

\textbf{Precision (PRE):} PRE measures the ratio of correctly predicted instances for a specific malware class to all predictions made for that class. It indicates how accurately the model predicts instances belonging to that particular class.

\begin{equation}
\text{PRE} = \frac{TP}{TP + FP}
\label{eq:precision}
\end{equation}

\textbf{Recall (REC):} REC calculates the ratio of correctly predicted instances for a specific malware class to all actual instances of that class in the dataset. It indicates the model's ability to find all relevant instances in that class.

\begin{equation}
\text{REC} = \frac{TP}{TP + FN}
\label{eq:recall}
\end{equation}

\textbf{F1-Score:} The F1-score provides a more balanced evaluation by incorporating both precision and recall. This enables a more accurate assessment of the model’s capacity to identify instances of the minority class, an essential consideration when dealing with imbalanced datasets. 

\begin{equation}
\text{F1-Score} = \frac{2 \cdot TP}{2 \cdot TP + FP + FN}
\label{eq:f1}
\end{equation}

\textbf{Area Under the Precision–Recall Curve (AUPRC):} The precision–recall curve characterizes the relationship between precision and recall across probability thresholds. AUPRC provides a threshold-independent measure of performance that is particularly informative for imbalanced datasets. In this study, it is computed for class $k$ in a one-vs-rest setting.

\begin{equation}
\text{AUPRC}_k
   = \int_0^1 \text{PRE}_k(\text{REC}_k) \, d\text{REC}_k
\end{equation}

\subsection{Performance Evaluation}

\subsubsection{Model Evaluation on LLM-Derived Embeddings}
Table \ref{table:performance_comparison_classif_models} compares the classification performance of the four models evaluated on LLM-generated behavioral embeddings. 
Our proposed model, BEACON, achieves the highest scores across all metrics, with accuracy, precision, recall, and F1-score all reaching 0.985, indicating a strong ability to extract discriminative features from the LLM-derived vectors.
Despite the use of different model architectures, all models achieve high performance. This observation suggests that the embeddings effectively capture meaningful behavioral signals across malware families. 
Specifically, the embeddings can capture patterns such as API usage, file access, and registry behavior in a dense vector format that retains both semantic and structural relationships essential for accurate classification. 
The consistently high F1 scores across all models further support the robustness of the embeddings under class imbalance, providing a strong foundation for behavior-based malware classification regardless of the subsequent model architecture.

\begin{table}[ht]
\caption{Performance of Classifiers on LLM-Derived Behavioral Embeddings}
\centering
\renewcommand{\arraystretch}{1.5} 
\begin{tabular}{|l|c|c|c|c|}
    \hline
    \textbf{Classification Model}  & \textbf{Acc.} & \textbf{Pre.} & \textbf{Rec.} & \textbf{F1-Score} \\ 
    \hline 
    SVM with RBF kernel & 0.976 & 0.977 & 0.959 & 0.966 \\
    \hline 
    MLP & 0.981 & 0.981 & 0.980 & 0.981 \\  
    \hline
    Bidirectional LSTM & 0.983 & 0.983 & 0.983 & 0.983 \\
    \hline
    \textbf{BEACON (Proposed)} & \textbf{0.985} & \textbf{0.985} & \textbf{0.985} & \textbf{0.985} \\
     \hline
\end{tabular}
\label{table:performance_comparison_classif_models}
\end{table}

Malware family prediction is treated as a multi-label classification task, which makes the evaluation results of our BEACON model per malware family particularly informative, as shown in Table~\ref{table:beacon_per_family_matrices}. The reported metrics provide a detailed breakdown of the model’s predictions across the ten malware families. As observed, misclassifications are minimal. Notably, even minority classes such as Adload (1.4\%) and HarHar (1.3\%) are correctly identified with high scores, with the latter achieving perfect scores of 1.0 across all four evaluation metrics. This observation is consistent with the overall high F1-scores and confirms the model's robustness to class imbalance.

\begin{table}[b]
\caption{Per-Family Classification Performance of BEACON on the Avast-CTU Dataset}
\centering
\renewcommand{\arraystretch}{1.5} 
\begin{tabular}{|l|c|c|c|c|}
    \hline
    \textbf{Malware family}  & \textbf{Acc.} & \textbf{Pre.} & \textbf{Rec.} & \textbf{F1-Score} \\ 
    \hline 
    Adload & 0.962 & 0.985  &   0.962  &   0.974 \\
    \hline 
    Emotet & 0.998 & 0.985   &  0.998   &  0.992 \\  
    \hline
    HarHar& 1.000 & 1.000  &   1.000   &  1.000\\
    \hline
    Lokibot& 0.962 & 0.966  &   0.962  &   0.964 \\
    \hline
    Qakbot& 0.988 & 0.999   &  0.988  &   0.994 \\
    \hline
    Swisyn& 0.999 & 0.998  &   0.999   &  0.998 \\
    \hline
    Trickbot& 0.980 & 0.990  &   0.980  &   0.985 \\
    \hline
    Ursnif& 0.950 & 0.967  &   0.950  &   0.959 \\
    \hline
    Zeus& 0.927 & 0.961  &   0.927  &   0.944 \\
    \hline
    njRAT& 0.966 & 0.952  &   0.966  &   0.959 \\
    \hline
\end{tabular}
\label{table:beacon_per_family_matrices}
\end{table}

\begin{figure}[h]
\centering
\includegraphics[width=0.96\linewidth]{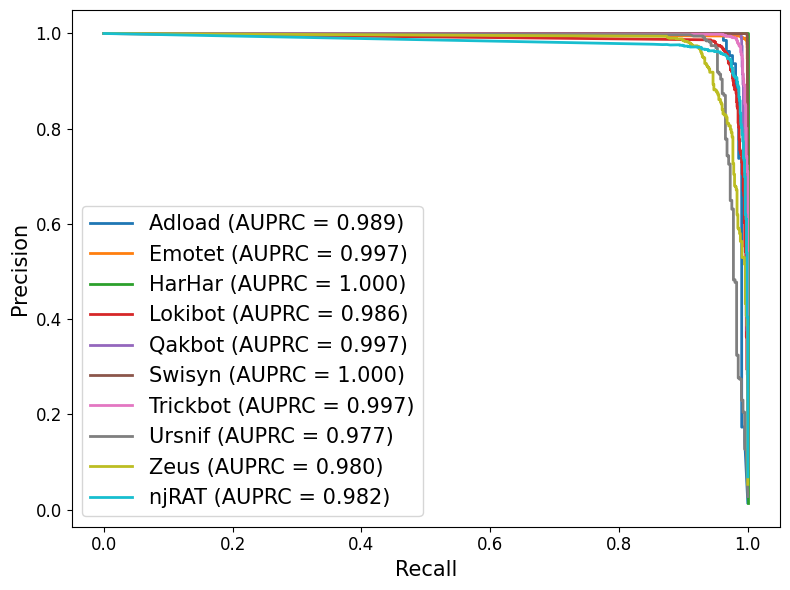}
\caption{One-vs-rest precision–recall curves for the ten malware families, with corresponding AUPRC values reported in the legend.}
\label{fig:pr_curves}
\end{figure}

The precision-recall curves for all ten malware families are illustrated in Figure \ref{fig:pr_curves}, with their corresponding AUPRC values in the legend. The consistently high AUPRC values confirm the effectiveness of the model in capturing class-specific patterns and maintaining strong precision–recall trade-offs. Notably, HarHar and Swisyn achieve perfect AUPRC scores of 1.000, suggesting that the model can identify these malware types without any false positives across the evaluated thresholds. Most other families, including Adload (0.989), Emotet (0.997), Qakbot (0.997), Trickbot (0.997), and njRAT (0.982), also reach near-perfect AUPRC values, demonstrating the robustness of the classifier in distinguishing malware samples. 
Ursnif (0.977) and Zeus (0.980) demonstrate slightly lower AUPRC values, suggesting that these families are marginally more difficult to separate, potentially due to behavioral similarities or overlapping features with other malware types.

\subsubsection{Comparison With Prior Work}
We directly compare the performance of our proposed framework with existing malware classification frameworks reported in the literature. Specifically, we evaluate both the F1-scores per malware family and the overall classification performance against prior studies that used the Avast-CTU Public CAPE dataset with the same train-test split, as summarized in Table~\ref{table:previous_work_per_family_f1_scores} and Table~\ref{table:previous_work_overall_performance}. 
To ensure comparability, we derived the accuracy, precision, recall, and F1-score values from the confusion matrices reported in \cite{bosansky2022avastdataset}.

Our proposed framework, BEACON, which integrates Gemini LLM-derived embeddings with the 1D CNN classifier,
achieves the highest test F1 scores in 8 out of 10 malware families on the Avast-CTU dataset (Table~\ref{table:previous_work_per_family_f1_scores}), with the remaining two families trailing by a minimal difference of just 0.001.
Notably, it maintains consistently strong performance across both frequent and infrequent classes. This suggests that BEACON effectively captures discriminative behavioral patterns even in minority classes, which often suffer from overfitting or neglect in conventional models. 
To isolate the impact of feature representation, we replicate the original MLP-based classification architecture from Bosansky et al.\cite{bosansky2022avastdataset}, substituting their inputs with our Gemini LLM-derived embeddings.
As shown in Table \ref{table:previous_work_overall_performance}, this substitution alone leads to improved performance over prior methods. Further performance gains in BEACON highlight the synergy between expressive feature representations and context-aware modeling of behavioral sequences.
In contrast to Nebula’s token-based approach, BEACON demonstrates consistent generalization across malware families with polymorphic characteristics (e.g., Emotet) and those with rich sub-variant structures (e.g., Zeus), while avoiding significant performance degradation in noisier or less frequent categories. This reinforces the importance of using deep contextual embeddings in malware behavior modeling and highlights the significance of tailoring deep learning pipelines to behavioral malware analysis.

\begin{table*}[ht]
\caption{Per-Family F1 Score Comparison With Prior Work on the Avast-CTU Dataset}
\centering
\renewcommand{\arraystretch}{1.5} 
\begin{tabular}{|l|c|c|c|c|c|c|c|c|c|c|}
    \hline
    & Adload & Emotet & HarHar & Lokibot & Qakbot & Swisyn & Trickbot & Ursnif & Zeus & njRAT \\ 
    \hline 
    Bosansky et al. \cite{bosansky2022avastdataset} & 0.825 &  0.927 & 0.993  & 0.960 &  \textbf{0.995} & \textbf{0.999}  &  0.953  &  0.891  & 0.482 & 0.947 \\
    \hline 
    Trizna et al. \cite{Trizna2024Nebula} & 0.439 & 0.939 & 0.776 & 0.896 & 0.988 & 0.997 & 0.923 & 0.936 & 0.642 & 0.866 \\  
    \hline
    Proposed BEACON & \textbf{0.974} & \textbf{0.992}  & \textbf{1.000} & \textbf{0.964} & 0.994 & 0.998  & \textbf{0.985} &  \textbf{0.959} & \textbf{0.944 }& \textbf{0.959} \\
    \hline
\end{tabular}
\label{table:previous_work_per_family_f1_scores}
\end{table*}

\begin{table*}[ht]
    \caption{Comparison of Overall Malware Classification Performance With Prior Work on the Avast-CTU Dataset}
    \centering
    \renewcommand{\arraystretch}{2} 
    \begin{tabular}{|c|c|c|c|c|c|c|c|c|}
        \hline
        \makecell{\textbf{}}
        & \makecell{\textbf{Feature} \\ \textbf{Representation}} 
        & \makecell{\textbf{Feature} \\ \textbf{Rep. Method}} 
        & \makecell{\textbf{Feature} \\ \textbf{Extraction Method}} 
        & \makecell{\textbf{Classification} \\ \textbf{Approach}} 
        & \textbf{Acc.} & \textbf{Pre.} & \textbf{Rec.} & \textbf{F1} \\ 
        \hline 
        Bosansky et al. \cite{bosansky2022avastdataset} & Semantic vectors & HMIL(JsonGrinder) &  \makecell{Schema inference \\ using JsonGrinder.jl} & MLP & 0.945 & 0.968 & 0.950 & 0.955 \\ 
        \hline 
        Trizna et al. \cite{Trizna2024Nebula} & Token sequence & Tokenization(BPE) & \makecell{Embedding and \\ positional encoding} & Transformer & - & - & - & 0.840 \\
        \hline 
        Our MLP & \makecell{Deep contextual \\ embeddings} & Gemini LLM & PCA & MLP & 0.981 & 0.981 & 0.980 & 0.981 \\
        \hline
        Proposed BEACON & \makecell{Deep contextual \\ embeddings} & Gemini LLM & PCA + CNN & CNN & \textbf{0.985} & \textbf{0.985} & \textbf{0.985} & \textbf{0.985} \\
        \hline
    \end{tabular} 
    \label{table:previous_work_overall_performance}
\end{table*}

\section{Conclusion}

In this study, we presented BEACON, a novel framework for malware classification that leverages dense contextual embeddings generated from raw behavioral reports using large language models (LLMs). This approach eliminates the need for manual feature engineering or hierarchical modeling, which are often required in traditional pipelines. By encoding complex malware behavior into structured vector representations, the embeddings capture both contextual and structural patterns, such as API usage, file activity, and registry operations, that are essential for distinguishing between malware families. 
Through extensive experiments with both traditional machine learning models (e.g., SVM) and deep learning models (e.g., MLP, LSTM, 1D CNN), we demonstrated that these LLM-derived embeddings enable accurate and robust classification, even under significant class imbalance. 
Among the models evaluated, our customized architecture, BEACON, achieved the highest performance across all metrics. 
Furthermore, comparative analysis with previous frameworks on the Avast-CTU dataset confirms that BEACON, combining Gemini LLM-derived embeddings with a tailored classification pipeline, consistently outperforms existing approaches in both per-family and overall classification performance.
These findings confirm the effectiveness of our model in extracting features from behavioral reports through dense embedding vectors, highlighting the value of combining LLM-based representations with deep learning for scalable and reliable behavior-based malware detection. 

As future work, we plan to develop an end-to-end framework with a custom embedding model tailored to malware behavior data. A domain-specific model can better capture low-level semantics, exploit structural patterns, and operate more efficiently than general-purpose LLMs, which makes it suitable for real-time or on-device detection. 
Additionally, we plan to incorporate explainable AI techniques to interpret learned representations and improve transparency in malware classification.

\bibliographystyle{IEEEtran}
\bibliography{ICMLA2025_LLM_Mal}

\end{document}